\documentclass[10pt,twocolumn,letterpaper]{article}

\usepackage{cvpr}
\usepackage{times}
\usepackage{epsfig}
\usepackage{graphicx}
\usepackage{amsmath}
\usepackage{amssymb}
\usepackage{multirow}
\usepackage{subfigure}
\usepackage{algorithmic}
\usepackage{amsthm}
\usepackage{booktabs}
\usepackage{graphicx,lipsum,afterpage}

\usepackage[ruled,lined]{algorithm2e}
\usepackage[bookmarks=false]{hyperref}

\cvprfinalcopy 

\def\cvprPaperID{5261} 

\ifcvprfinal\fi
\begin{document}

\title{Revisiting Knowledge Distillation via Label Smoothing Regularization}

\author{Li Yuan\textsuperscript{\rm 1}\quad \,  Francis EH Tay\textsuperscript{\rm 1}\quad \, Guilin Li\textsuperscript{\rm 2} \quad \, Tao Wang\textsuperscript{\rm 1} \quad \, Jiashi Feng\textsuperscript{\rm 1}\\\\
\textsuperscript{\rm 1}National University of Singapore \quad \textsuperscript{\rm 2}Huawei Noah's Ark Lab\\
{\{ylustcnus, twangnh\}@gmail.com},\,
 {\{mpetayeh,elefjia\}@nus.edu.sg},\,{guilinli2@huawei.com}
}

\maketitle
\thispagestyle{empty}

\begin{abstract}
\vspace{-10pt}
Knowledge Distillation (KD) aims to distill the knowledge of a cumbersome teacher model into a lightweight student model. Its success is generally attributed to the privileged information on similarities among categories provided by the teacher model, and in this sense, only strong
teacher models are deployed to teach weaker students in
practice. In this work, we challenge this common belief
by following experimental observations: 1) beyond the acknowledgment that the teacher can improve the student, the
student can also enhance the teacher significantly by reversing the KD procedure; 2) a poorly-trained teacher with
much lower accuracy than the student can still improve the
latter significantly. To explain these observations, we provide a theoretical analysis of the relationships between KD
and label smoothing regularization. We prove that 1) KD
is a type of learned label smoothing regularization and 2)
label smoothing regularization provides a virtual teacher
model for KD. From these results, we argue that the success
of KD is not fully due to the similarity information between
categories from teachers, but also to the regularization of
soft targets, which is equally or even more important.

Based on these analyses, we further propose a novel
Teacher-free Knowledge Distillation (Tf-KD) framework,
where a student model learns from itself or manuallydesigned regularization distribution. The Tf-KD achieves
comparable performance with normal KD from a superior teacher, which is well applied when a stronger teacher
model is unavailable. Meanwhile, Tf-KD is generic and
can be directly deployed for training deep neural networks.
Without any extra computation cost, Tf-KD achieves up
to 0.65\% improvement on ImageNet over well-established
baseline models, which is superior to label smoothing regularization. 

\end{abstract}
\vspace{-4mm}
\section{Introduction}
Knowledge Distillation (KD)~\cite{hinton2015distilling} aims to transfer knowledge from one neural network (teacher) to another (student). Usually, the teacher model has a strong learning capacity with higher performance, which teaches a lower-capacity student model through providing ``soft targets''. It is commonly believed that the soft targets of the teacher model can transfer ``dark knowledge'' containing privileged information on similarity among different categories~\cite{hinton2015distilling} to enhance the student model.

In this work, we first examine such a common belief through following exploratory experiments:
1) let student models teach teacher models by transferring soft targets of the students; (2) let poorly-trained teacher models with worse performance teach students. 
Based on the common belief, it is expected that the teacher model would not be enhanced significantly via training from the students and poorly-trained teachers would not enhance the students, as the weak student and poorly-trained teacher models cannot provide reliable similarity information between categories. However, after extensive experiments on various models and datasets, we 
observe contradictory results: the weak student can improve the teacher and the poorly-trained teacher can also enhance the student remarkably. 
Such intriguing results motivate us to interpret KD as a regularization term, and we re-examine knowledge distillation from the perspective of Label Smoothing Regularization (LSR)~\cite{szegedy2016rethinking}
that regularizes model training by replacing the one-hot labels with smoothed ones. \footnote{Code: https://github.com/yuanli2333/Teacher-free-Knowledge-Distillation}

We then analyze theoretically the relationships between KD and LSR. For LSR, by splitting the smoothed label into two parts and examining the corresponding losses, we find the first part is the ordinary cross-entropy for ground-truth distribution (one-hot label) and outputs of model, and the second part corresponds to a virtual teacher model which provides a uniform distribution to teach the model. For KD, by combining the teacher's soft targets with the one-hot ground-truth label, we find that KD is a learned LSR where the smoothing distribution of KD is from a teacher model but the smoothing distribution of LSR is manually designed. In a nutshell, we find \textit{KD is a learned LSR and LSR is an ad-hoc KD}. Such relationships can explain the above counterintuitive results\textemdash the soft targets from weak student and poorly-trained teacher models can effectively regularize the model training, even though they lack strong similarity information between categories. We therefore argue that the similarity information between categories cannot fully explain the dark knowledge in KD, and the soft targets from the teacher model indeed provide effective regularization for the student model, which are equally or even more important.

Based on the analyses, we conjecture that with non-reliable or even zero similarity information between categories from the teacher model, KD may still well improve the student models. 
We thus propose a novel Teacher-free Knowledge Distillation (Tf-KD) framework with two implementations. The first one is to train the student model by itself (i.e., self-training), and the second is to manually design a target distribution as a virtual teacher model which has 100\% accuracy. The first method is motivated by replacing the dark knowledge with predictions from the model itself, and  
the second method is inspired by the relationships between KD and LSR.  
We validate through extensive experiments that the two implementations of Tf-KD are both simple yet effective. Particularly, in the second implementation without similarity information in the virtual teacher, Tf-KD still achieves comparable performance with normal KD, which clearly justifies:  
\begin{center}
    \textit{Dark knowledge does not just include the similarity between categories, but also imposes  regularization on the student training}. 
\end{center}
Tf-KD well applies to scenarios where the student model is too strong to find teacher models or computational resource is limited for training teacher models. For example, if we take a cumbersome single model ResNeXt101-32$\times$8d~\cite{xie2017aggregated} as the student model (with 88.79M parameters and 16.51G FLOPs on ImageNet), it is hard or computationally expensive to train a stronger teacher model. We deploy our virtual teacher to teach this powerful student and achieve 0.48\% improvement on ImageNet without any extra computation cost. Similarly, when taking a powerful single model ResNeXt29-8$\times$64d with 34.53M parameters as a student model, our self-training implementation achieves more than 1.0\% improvement on CIFAR100 (from 81.03\% to 82.08\%).

Our contributions are summarized as follows:
\begin{itemize}
\item By designing two exploratory experiments on teacher models of KD, we observe counterintuitive results, which motivate us to interpreted KD as a regularization method.  
\vspace{-3pt}
\item We then provide theoretical analysis to reveal the relationships between KD and label smoothing regularization.
\vspace{-3pt}
\item We propose Teacher-free Knowledge Distillation (Tf-KD), which achieves comparable performance with normal knowledge distillation and superior performance to label smoothing regularization on ImageNet-2012. 
\end{itemize}

\section{Exploratory Experiments and Counterintuitive Observations}
\label{sec: tinking experiments}

To examine the common belief on dark knowledge in KD, we conduct two exploratory experiments:

\begin{enumerate}
\item[1)] The standard knowledge distillation is to adopt a teacher to teach a weaker student. What if we reverse the operation? Based on the common belief, the teacher should not be improved significantly because the student is too weak to transfer effective knowledge. 

\item[2)] If we use a poorly-trained teacher which has much worse performance than the student to teach the student, it is assumed to bring no improvement to the latter. For example, if a poorly-trained teacher with only 10\% accuracy is adopted in an image classification task, the student would learn from its soft targets with 90\% error, thus the student should not be improved or even suffer worse performance.
\end{enumerate}
\begin{figure*}[!h]
\begin{center}
\subfigure[Normal KD]{
\includegraphics[scale=0.35]{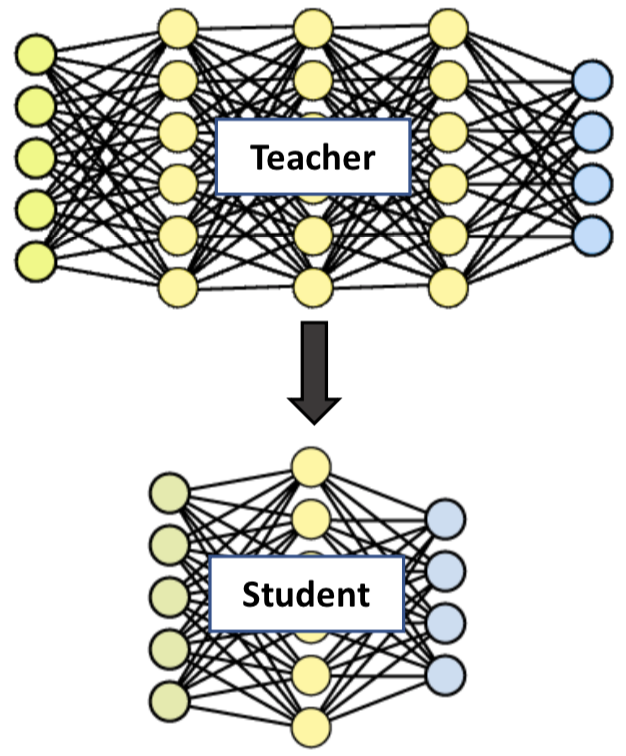}
}
\subfigure[Reversed KD]{
\includegraphics[scale=0.35]{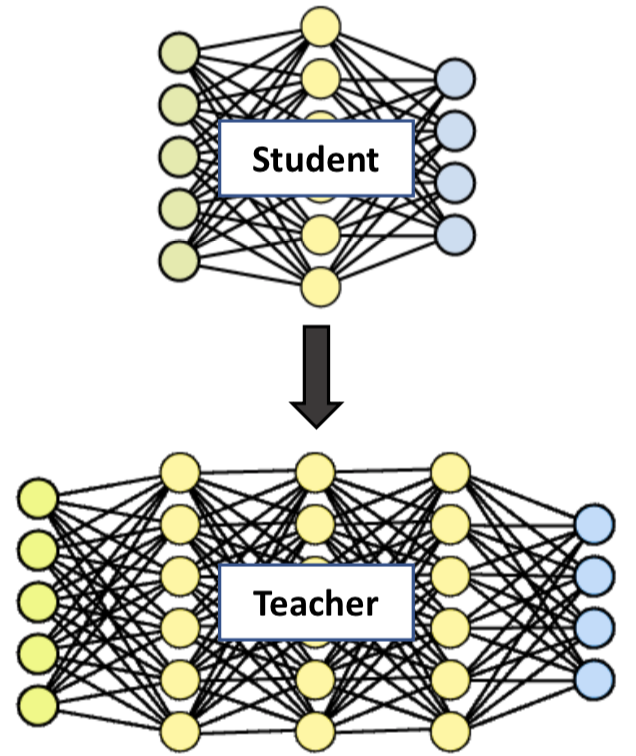}
}
\subfigure[Defective KD]{
\includegraphics[scale=0.35]{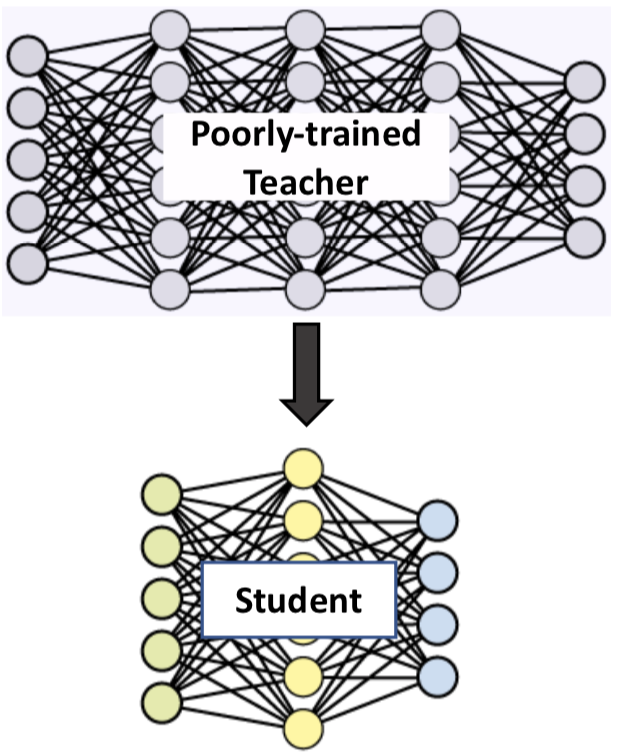}
}
\caption{(a) Normal KD framework. (b)(c) Diagrams of exploratory experiments we conduct.}
\label{fig:exploratory experiments}
\vspace{-20pt}
\end{center}
\end{figure*}

We name the ``student teach teacher'' as Reversed Knowledge Distillation (Re-KD), and the ``poorly-trained teacher teach student'' as Defective  Knowledge Distillation (De-KD) (Fig.~\ref{fig:exploratory experiments}). We conduct Re-KD and De-KD experiments on CIFAR10, CIFAR100 and Tiny-ImageNet datasets with a variety of neural networks. For fair comparisons, all experiments are conducted with the same settings and hyper-parameters are obtained by grid search from 70 epochs training (200 epochs in total). Detailed implementation and experiment settings are given in Supplementary Material.

\subsection{Reversed Knowledge Distillation}
We conduct Re-KD experiments on the three datasets respectively. CIFAR10 and CIFAR100~\cite{krizhevsky2009learning} contain natural RGB images of 32x32 pixels with 10 and 100 classes, respectively, and Tiny-ImageNet is a subset of ImageNet~\cite{deng2009imagenet} with 200 classes, where each image is down-sized to 64x64 pixels. For generality of the experiments, we adopt 5-layer plain CNN, MobilenetV2~\cite{sandler2018mobilenetv2} and ShufflenetV2~\cite{ma2018shufflenet} as student models and ResNet18, ResNet50~\cite{he2016deep}, DenseNet121~\cite{huang2017densely} and ResNeXt29-8$\times$64d as teachers. The results of Re-KD on the three datasets are given in Tabs.~\ref{tab: Re-KD_CIFAR100} to~\ref{tab:Re-KD_tiny}.

In Tab.~\ref{tab: Re-KD_CIFAR100}, the teacher models are improved significantly by learning from students, especially for teacher models ResNet18 and ResNet50. The two teachers obtain more than 1.1\% improvement when taught by MobileNetV2 and ShuffleNetV2. We can also observe similar results on CIFAR10 and Tiny-ImageNet. 
When comparing Re-KD (S$\rightarrow$T) with Normal KD (T$\rightarrow$S), we can see in most cases, Normal KD achieves better results. It should be noted that Re-KD takes the teacher's accuracy as the baseline accuracy, which is much higher than that of Normal KD. However, in some cases, we can find Re-KD outperforms Normal KD. For instance, in Tab.~\ref{tab: Re-KD_cifar10} (3rd row), the student model (plain CNN) can only be improved by 0.31\% when taught by MobileNetV2, but the teacher (MobileNetV2) can be improved by 0.92\% by learning from the student. We have similar observations for ResNeXt29 and ResNet18 (4th row in Tab.~\ref{tab: Re-KD_cifar10}).

\begin{table*}[!h]
\centering
\renewcommand\arraystretch{1.0}
\small
\caption{Normal KD and Re-KD experiment results on CIFAR100.  We report mean$\pm$std (in $\%$) over 3 runs. The number in parenthesis means increased accuracy over baseline (T: teacher, S: student).} 
\label{tab: Re-KD_CIFAR100}
\begin{tabular}{l|l|c|c}
\toprule
Teacher: baseline &Student: baseline  &Normal KD (T$\rightarrow$S)   &Re-KD (S$\rightarrow$T)       \\ \midrule
\multirow{2}{*}{ResNet18: 75.87} & MobileNetV2: 68.38   & 71.05$\pm$0.16 (\textbf{+2.67}) & 77.28$\pm$0.28 (\textbf{+1.41)} \\ 
                          & ShuffleNetV2: 70.34  & 72.05$\pm$0.13 (\textbf{+1.71})  &77.35$\pm$0.32 (\textbf{+1.48})    \\ \midrule
                          
\multirow{2}{*}{ResNet50: 78.16} & MobileNetV2: 68.38  &71.04$\pm$0.20 (\textbf{+2.66}) &79.30$\pm$0.11 (\textbf{+1.14}) \\ 
                           & ShuffleNetV2: 70.34 &72.15$\pm$0.18  (\textbf{+1.81}) &79.43$\pm$0.39 (\textbf{+1.27}) \\ \midrule
                           
\multirow{2}{*}{DenseNet121: 79.04} & MobileNetV2: 68.38  &71.29$\pm$0.23 (\textbf{+2.91}) &79.55$\pm$0.11 (\textbf{+0.51})\\ 
                           & ShuffleNetV2: 70.34 &72.32$\pm$0.25 (\textbf{+1.98})& 79.83$\pm$0.05 (\textbf{+0.79})  \\ \midrule 
                           
\multirow{2}{*}{ResNeXt29: 81.03}  & MobileNetV2: 68.38 &71.65$\pm$0.41 (\textbf{+3.27}) &81.53$\pm$0.14 (\textbf{+0.50})  \\ 
                           & ResNet18:\qquad 75.87   & 77.84$\pm$0.15 (\textbf{+1.97})  & 81.62$\pm$0.22 (\textbf{+0.59})
                           \\ \bottomrule
\end{tabular}
\vspace{-5pt}
\end{table*}

\begin{table*}[!h]
\centering
\renewcommand\arraystretch{1.0}
\small
\caption{Re-KD experiment results (accuracy, mean$\pm$std over 3 runs in \%) on CIFAR10.}
\label{tab: Re-KD_cifar10}
\begin{tabular}{l|l|c|c}
\toprule
 Teacher: baseline & Student: baseline   & Normal KD (T$\rightarrow$S) & Re-KD (S$\rightarrow$T) \\ \midrule
\multirow{2}{*}{ResNet18: 95.12} & Plain CNN: 87.14 & 87.67$\pm$0.17 (\textbf{+0.53}) & 95.33$\pm$0.12 \textbf{(+0.21)} \\ 
                          & MobileNetV2: 90.98  &91.69$\pm$0.14 \textbf{(+0.71)} & 95.71$\pm$0.11 \textbf{(+0.59)}         \\ \midrule
                           MobileNetV2: 90.98 & Plain CNN: 87.14 &87.45$\pm$0.18 (\textbf{+0.31}) & 91.81$\pm$0.23 (\textbf{+0.92})   \\ \midrule
                         ResNeXt29: 95.76 & ResNet18: 95.12    &95.80$\pm$0.13 (\textbf{+0.68}) & 96.49$\pm$0.15 (\textbf{+0.73})    \\ \bottomrule
\end{tabular}
\vspace{-5pt}
\end{table*}

\begin{table*}[!h]
\centering
\renewcommand\arraystretch{1.0}
\small
\caption{Re-KD experiment results (accuracy, in $\%$) on Tiny-ImageNet.}
\label{tab:Re-KD_tiny}
\begin{tabular}{l|l|c|c}
\toprule
Teacher: baseline & Student: baseline & Normal KD (T$\rightarrow$S)  & Re-KD (S$\rightarrow$T)   \\ \midrule
\multirow{2}{*}{ResNet18: 63.44} & MobileNetV2: 55.06 & 56.70 (\textbf{+1.64}) & 64.12 (\textbf{+0.68)} \\ 
            & ShuffleNetV2: 60.51  &61.19 (\textbf{+0.68}) &64.35 (\textbf{+0.91})\\ \midrule
\multirow{3}{*}{ResNet50: 67.47} & MobileNetV2: 55.06 &56.02 (\textbf{+0.96}) & 67.68 (\textbf{+0.21}) \\ 
            & ShuffleNetV2: 60.51 & 60.79 (\textbf{+0.28}) &67.62 (\textbf{+0.15}) \\ 
             & ResNet18:\qquad63.44  & 64.23 (\textbf{+0.79}) & 67.89 (\textbf{+0.42}) \\  \bottomrule
\end{tabular}
\label{-10pt}
\end{table*}

We claim that while the standard knowledge distillation can improve the performance of students on all datasets, the superior teacher can also be enhanced significantly by learning from a weak student, as suggested through the Re-KD experiments.

\subsection{Defective Knowledge Distillation}
We conduct De-KD on CIFAR100 and Tiny-ImageNet. We adopt MobileNetV2 and ShuffleNetV2 as student models and ResNet18, ResNet50 and ResNeXt29 (8$\times$64d) as teacher models. The poorly-trained teachers are trained by 1 epoch (ResNet18) or 50 epochs (ResNet50 and ResNeXt29), with very poor performance. For example, ResNet18 only obtains 15.48\% accuracy on CIFAR100 and 9.41\% accuracy on Tiny-ImageNet after trained with 1 epoch, and ResNet50 obtains 45.82\% and 31.01\% on CIFAR100 and Tiny-ImageNet, after trained with 50 epochs (200 epochs in total).

From De-KD experiment results on CIFAR100 in Tab.~\ref{tab: De-KD}, we observe that the student can be greatly promoted even when distilled by a poorly-trained teacher. For instance, the MobileNetV2 and ShuffleNetV2 can be promoted by 2.27\% and 1.48\% when taught by the one-epoch-trained ResNet18 with only 15.48\% accuracy (2nd row). For poorly-trained ResNeXt29 with 51.94\% accuracy (4th row), we find ResNet18 can still be improved by 1.41\%, and MobileNetV2 obtains 3.14\% improvement. From the De-KD experiment results on Tiny-ImageNet in Tab.~\ref{tab: De-KD}, we find ResNet18 with 9.14\% accuracy can still enhance the teacher model MobileNetV2 by 1.16\%. Other poorly-trained teachers are all able to enhance the students to some degree.

\begin{figure*}[!h]
\begin{center}
\subfigure[ResNet18]{
\label{fig: mv2_kd_r18}
\includegraphics[scale=0.35]{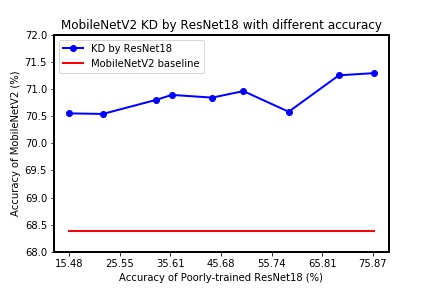}
}
\subfigure[ResNeXt29]{
\label{fig: mv2_kd_rx29}
\includegraphics[scale=0.35]{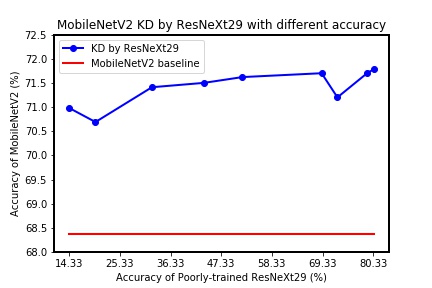}
}
\caption{MobileNetV2 taught by ResNet18 and ResNeXt29 with different accuracy on CIFAR100. MobileNetV2 is enhanced by different poorly-trained teachers compared with baseline (the red line). The final point of two blue lines is the result taught by ``fully-trained teacher''.}
\label{fig: mv2_kd_r18_rx29}
\vspace{-20pt}
\end{center}
\end{figure*}

To better demonstrate the distillation accuracy of a student when taught by poorly-trained teachers with different levels of accuracy, we save 9 checkpoints of ResNet18 and ResNeXt29 in the normal training process. Taking these checkpoints as teacher models to teach MobileNetV2, we observe that MobileNetV2 can always be improved by poorly-trained ResNet18 or poorly-trained ResNeXt29 with different levels of accuracy (Fig.~\ref{fig: mv2_kd_r18_rx29}). So we can say while a poorly-trained teacher provides much more noisy logits to the student, the student can still be enhanced. The De-KD experiment results are also conflicted with the common belief.

The counterintuitive results of Re-KD and De-KD make us rethink the ``dark knowledge'' in KD, and we argue that it does not just contain the similarity information. Lacking enough similarity information, a model can still provide ``dark knowledge'' to enhance other models.  To explain this, we make a reasonable assumption and view knowledge distillation as a model regularization, and investigate what is the additional information in the ``dark knowledge'' of a model. In the next, we will analyze the relationships between knowledge distillation and label smoothing regularization to explain the experimental results of Re-KD and De-KD. 

\begin{table*}[!h]
\centering
\renewcommand\arraystretch{1.0}
\small
\caption{De-KD accuracy (in \%) on two datasets. Pt-Teacher is ``Poorly-trained Teacher''. Refer to the ``Normal KD'' in Tabs.~\ref{tab: Re-KD_CIFAR100} to~\ref{tab:Re-KD_tiny} for the accuracy of students taught by ``fully-trained teacher''.}
\label{tab: De-KD}
\begin{tabular}{l|l|l|c}
\toprule
Dataset & Pt-Teacher: baseline & Student: baseline &De-KD  \\ \midrule
\multirow{8}{*}{CIFAR100} & \multirow{2}{*}{ResNet18: 15.48} & MobileNetV2: 68.38  &70.65$\pm$0.35 (\textbf{+2.27}) \\ 
                           && ShuffleNetV2: 70.34 & 71.82$\pm$0.11 (\textbf{+1.48}) \\ \cmidrule{2-4}  
                           
            & \multirow{3}{*}{ResNet50: 45.82} & MobileNetV2: 68.38  &71.45$\pm$0.23 (\textbf{+3.09}) \\ 
                           & & ShuffleNetV2: 70.34 &72.11$\pm$0.09 (\textbf{+1.77}) \\ 
                           & & ResNet18: 75.87 &77.23$\pm$0.11 (\textbf{+1.23}) \\ \cmidrule{2-4}
            & \multirow{3}{*}{ResNeXt29: 51.94} & MobileNetV2: 68.38  &71.52$\pm$0.27 (\textbf{+3.14}) \\ 
                           && ShuffleNetV2:70.34 &72.26$\pm$0.36 (\textbf{+1.92}) \\ 
                           && ResNet18: 75.87 &77.28$\pm$0.17 (\textbf{+1.41}) \\ \midrule 

\multirow{4}{*}{Tiny-ImageNet} & \multirow{2}{*}{ResNet18: 9.41} & MobileNetV2: 55.06  &56.22 (\textbf{+1.16}) \\ 
        & & ShuffleNetV2: 60.51 & 60.66 (\textbf{+0.15})\\  \cmidrule{2-4}
        
        
        & \multirow{2}{*}{ResNet50: 31.01} & MobileNetV2:55.06  &56.02 (\textbf{+0.96}) \\ 
        & & ShuffleNetV2: 60.51 &61.09 (\textbf{+0.58})\\ \bottomrule
\end{tabular}
\vspace{-10pt}
\end{table*}

\section{Knowledge Distillation and Label Smoothing Regularization}
\label{sec: KD and LS}
We mathematically analyze the relationships between Knowledge Distillation (KD) and Label Smoothing Regularization (LSR), hoping to explain the intriguing results of exploratory experiments in Sec.~\ref{sec: tinking experiments}. Given a neural network $S$ to train, we first give loss function of LSR for $S$. For each training example $x$, $S$ outputs the probability of each label $k\in\left \{ 1...K \right \}: p(k|x)=softmax(z_{k})=\frac{\exp(z_{k})}{\sum_{i=1}^{K}\exp(z_{i})}$, where $z_{i}$ is the logit of the neural network $S$. The ground truth distribution over the labels is $q(k|x)$. We write $p(k|x)$ as $p(k)$ and $q(k|x)$ as $q(k)$ for simplicity. The model $S$ can be trained by minimizing the cross-entropy loss: $H(q,p)=-\sum_{k=1}^{K}q(k)\log(p(k))$. For a single ground-truth label $y$, the $q(y|x)=1$ and $q(k|x)=0$ for all $k\neq y$. 

In LSR, it minimizes the cross-entropy between modified label distribution ${q}'(k)$ and the network output $p(k)$, where ${q}'(k)$ is the smoothed label distribution formulated as
\vspace{-5pt}
\begin{equation}
{q}'(k)=(1-\alpha)q(k) + \alpha u(k),
\label{eqn:LS_q}
\end{equation}
which is a mixture of $q(k)$ and a fixed distribution $u(k)$, with weight $\alpha$. Usually, the $u(k)$ is uniform distribution as $u(k)=1/K$. The cross-entropy loss $H({q}', p)$ defined over the smoothed labels is
\vspace{-5pt}
\begin{equation}
\label{eqn:CE_loss}
\begin{aligned}
    H({q}', p)&=-\sum_{k=1}^{K}{q}'(k)\log p(k)=(1-\alpha)H(q,p)+\alpha H(u, p) \\
              &=(1-\alpha)H(q,p)+\alpha (D_{KL}(u, p)+H(u)),
\end{aligned}
\end{equation}
where $D_{KL}$ is the Kullback-Leibler divergence (KL divergence) and $H(u)$ denotes the entropy of $u$ and is a constant for the fixed uniform distribution $u(k)$. Thus, the loss function of label smoothing to model $S$ can be written as
\begin{equation}
\label{eqn:LS_loss}
    \mathcal{L}_{LS}=(1-\alpha)H(q,p)+\alpha D_{KL}(u, p).
\end{equation}

For knowledge distillation, the teacher-student learning mechanism is applied to improve the performance of the student. We assume the student is the model $S$ with output prediction $p(k)$, and the output prediction of the teacher network is $p^{t}_{\tau}(k)=softmax(z^{t}_{k})=\frac{\exp(z^{t}_{k}/\tau)}{\sum_{i=1}^{K}\exp(z^{t}_{i}/\tau)}$, where $z^{t}$ is the output logits of the teacher network and $\tau$ is the temperature to soften $p^{t}(k)$ (written as $p^{t}_{\tau}(k)$ after softened). The idea behind knowledge distillation is to let the student (the model $S$) mimic the teacher by minimizing the cross-entropy loss and KL divergence between the predictions of student and teacher as
\begin{equation}
\label{eqn:KD_loss}
    \mathcal{L}_{KD} = (1-\alpha)H(q, p) + \alpha D_{KL}(p^{t}_{\tau}, p_{\tau}).
\end{equation}

Comparing Eq.~(\ref{eqn:LS_loss}) and Eq.~(\ref{eqn:KD_loss}), we find the two loss functions have a similar form.  The only difference is that the $p^{t}_{\tau}(k)$ in $D_{KL}(p^{t}_{\tau}, p_{\tau})$ is a distribution from a teacher model and $u(k)$ in $D_{KL}(u, p)$ is the pre-defined uniform distribution. From this view, we can consider KD as a special case of LSR where the smoothing distribution is learned but not pre-defined. On the other hand, if we view the regularization term $D_{KL}(u, p)$ as a virtual teacher model of knowledge distillation, this teacher model will give a uniform probability to all classes, meaning it has a random accuracy (1\% accuracy for CIFAR100, 0.1\% accuracy for ImageNet).

Since $ D_{KL}(p^{t}_{\tau}, p_{\tau})=H(p^{t}_{\tau}, p_{\tau})-H(p^{t}_{\tau})$, where the entropy $H(p^{t}_{\tau})$ is constant for a  fixed  teacher model, we can reformulate Eq.~(\ref{eqn:KD_loss}) to
\begin{equation}
\begin{aligned}
    L_{KD} 
    & = (1-\alpha)H(q, p) + \alpha (D_{KL}(p^{t}_{\tau}, p_{\tau})+ H(p^{t}_{\tau}))\\
    &= (1-\alpha)H(q, p) + \alpha H(p^{t}_{\tau}, p_{\tau}).
\end{aligned}
\end{equation}
If we set the temperature $\tau=1$, we have $L_{KD} = H(\tilde{q}^{t}, p)$, where $\tilde{q}^{t}$ is
\begin{equation}
\tilde{q}^{t}(k)=(1-\alpha)q(k) + \alpha p^{t}(k).
\label{eqn:KD_q}
\end{equation}
If we compare Eq.~(\ref{eqn:KD_q}) with Eq.~(\ref{eqn:LS_q}), it is more clearly seen that KD is a special case of LSR. Moreover, the distribution $p^{t}(k)$ is a learned distribution (from a trained teacher) instead of a uniform distribution $u(k)$. We visualize the output probability $p^{t}(k)$ of a teacher and compare it with label smoothing in Supplementary Material, and find with higher temperature $\tau$, the $p^{t}(k)$ is more similar to the uniform distribution $u(k)$ of label smoothing.

Based on the comparison of the two loss functions, we summarize the relationships between knowledge distillation and label smoothing regularization as follows:
\begin{itemize}
\item Knowledge distillation is a learned label smoothing regularization, which has a similar function with the latter, i.e.\ regularizing the classifier layer of the model.
\item Label smoothing is an ad-hoc knowledge distillation, which can be revisited as a teacher model with random accuracy and temperature $\tau=1$.
\item With higher temperature, the distribution of teacher's soft targets in knowledge distillation is more similar to the uniform distribution of label smoothing.
\end{itemize}

Therefore, the experiment results of Re-KD and De-KD can be explained as the soft targets of the model in high temperature are closer to a uniform distribution of label smoothing, where the learned soft targets can provide model regularization for the teacher model. That is why a student can enhance the teacher and a poorly-trained teacher can still improve the student model. 


\section{Teacher-free Knowledge Distillation}
As we above analyzed, the ``dark knowledge'' in the teacher model is more of a regularization term than the similarity information between categories. Intuitively, we consider replacing the output distribution of the teacher model with a simple one. We therefore propose a novel Teacher-free Knowledge Distillation (Tf-KD) framework with two implementations. Tf-KD is especially applicable to cases where a stronger teacher model is not available, or only limited computation resources are provided. 

The first Tf-KD method is self-training knowledge distillation, denoted as Tf-KD$_{self}$. As aforementioned,  the teacher can be taught by a student and a poorly-trained teacher can also enhance the student. Hence when a stronger teacher model is not available, we propose to deploy ``self-training''. It should be noted that the teacher in KD always means a stronger model. We name self-training as a teacher-free method because the model is not a teacher with stronger learning capacity than itself. Our Tf-KD$_{self}$ is similar to Born-again networks~\cite{furlanello2018born}, but there are two differences. Our motivation (self-training/self-regularization) is different from Born-again networks; and our method use soft targets of model self as regularization, while Born-again networks utilize an ensemble of student models to train itself iteratively. Specifically, we first train the student model in the normal way to obtain a pre-trained model, which is then used to provide soft label to train itself as in Eq.~(\ref{eqn:KD_loss}). Formally, given a model $S$, we denote its pre-trained model as $S^{p}$; then we try to minimize the KL divergence of the logits between $S$ and $S^{p}$ by Tf-KD$_{self}$. The loss function of Tf-KD$_{self}$ to train model $S$ is
\begin{equation}
\label{eqn:Tf-KD_self}
    L_{self} = (1-\alpha)H(q, p) + \alpha D_{KL}(p^{t}_{\tau}, p_{\tau}),
\end{equation}
where $p$, $p^{t}_{\tau}$ are the output probability of $S$ and $S^{p}$ respectively, $\tau$ is the temperature and $\alpha$ is the weight.

The second implementation of our Tf-KD method is to manually design a teacher with 100\% accuracy. In Sec.~\ref{sec: KD and LS}, we reveal LSR is a virtual teacher model with random accuracy. So, if we design a teacher with higher accuracy, we can assume it would bring more improvement to the student. We propose to combine KD and LSR to build a simple teacher model which will output distribution for classes as the following:
\begin{equation}
p^{d}(k)=\begin{cases}
a & \text{ if } k=c, \\ 
(1-a)/(K-1) & \text{ if } k\neq c, 
\end{cases}
\label{eqn:T_probability}
\end{equation}
where $K$ is the total number of classes, $c$ is the correct label and $a$ is the correct probability for the correct class. We always set $a\geq 0.9$, so the probability of a correct class is much higher than that of an incorrect one, and the manually-designed teacher model has 100\% accuracy for any dataset.
\begin{figure}
\begin{center}
\includegraphics[scale=0.38]{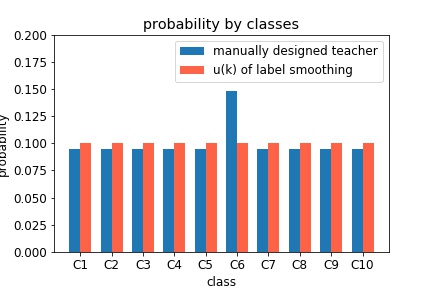}
\end{center}
\caption{Distribution of manually designed teacher (softened by $\tau=20$) on 10-class dataset. C6 is the correct label. As a comparison, the orange bar is the uniform distribution of LSR.}
\vspace{-15pt}
\label{fig:designe-teacher}
\end{figure}
We name this method as Teacher-free KD by manually-designed regularization, denoted as Tf-KD$_{reg}$. The loss function is
\begin{equation}
\label{eqn:Tf-KD_reg}
    L_{reg} = (1-\alpha)H(q, p) + \alpha D_{KL}(p^{d}_{\tau}, p_{\tau}),
\end{equation}
where $\tau$ is the temperature to soften the manually-designed distribution $p^{d}$ (as $p^{d}_{\tau}$ after softening). We set a high temperature $\tau\geq 20$ to make this virtual teacher output a soft probability, in which way it gains the smoothing property as LSR.
We visualize the distribution of the manually designed teacher in Fig.~\ref{fig:designe-teacher}.  As Fig.~~\ref{fig:designe-teacher} shows, this manually designed teacher model outputs soft targets with 100\% classification accuracy, and also has the smoothing property of label smoothing. But the Tf-KD$_{reg}$ is not an over-parameterized version of LSR because the temperature $\tau\gg1$, thus Eq.~\ref{eqn:Tf-KD_reg} will not be equal to Eq.~\ref{eqn:LS_loss} when we adjust the parameters $\alpha$, $a$ or $u(k)$.

The two Teacher-free methods, Tf-KD$_{self}$ and Tf-KD$_{reg}$, are very simple yet effective, as validated via extensive experiments in the next section.

\section{Experiments on Tf-KD}
In this section, we conduct experiments to evaluate Tf-KD$_{self}$ and Tf-KD$_{reg}$ on three datasets for image classification:  CIFAR100, Tiny-ImageNet and ImageNet. For fair comparisons, all experiments are conducted with the same setting.
\vspace{-4pt}
\subsection{Experiments for Self-training}
For our Tf-KD$_{self}$ and Normal KD, the hyper-parameters (temperature $\tau$ and $\alpha$) are obtained by grid search from 70 epochs training (200 epochs), the values of hyper-parameters are given in Supplementary Material.
\vspace{-7pt}
\paragraph{CIFAR100.} On CIFAR100, we use baseline models including MobileNetV2, ShuffleNetV2, GoogLeNet,
ResNet18, DenseNet121 and ResNeXt29(8$\times$64d). The baselines are trained for 200 epochs, with batch size 128. The initial learning rate is 0.1 and then divided by 5 at the 60th, 120th, 160th epoch. We use SGD optimizer with the momentum of 0.9, and weight decay is set to 5e-4. 

\begin{figure}[h]
\centering 
\includegraphics[width=7.5cm]{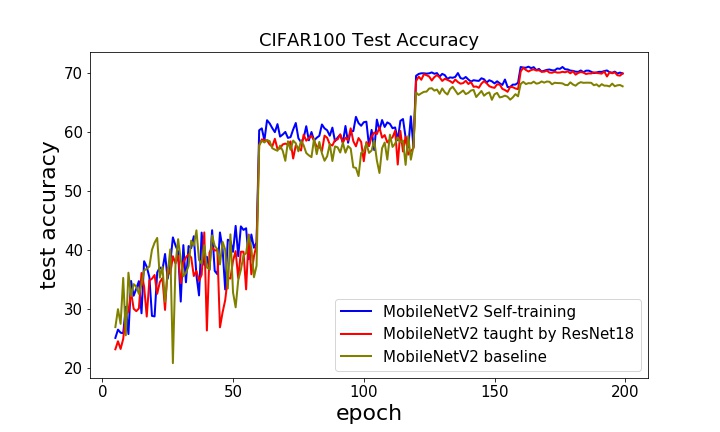}
\caption{MobileNetV2 obtains similar improvement by self-regularization or taught by ResNet18.}
\label{fig: self_KD_N_KD}
\vspace{-23pt}
\end{figure}

Tab.~\ref{tab:self_CIFAR100} shows the test accuracy of the six models. It can be seen that our Tf-KD$_{self}$ consistently outperforms the baselines. For example, as a powerful model with 34.52M parameters, ResNeXt29 improves itself by 1.05\% with self-regularization. Even when compared to Normal KD with a superior teacher in Tab.~\ref{tab:self_CIFAR100} (4th column), our method achieves comparable performance (experiment settings for Tf-KD and Normal KD are the same and hyper-parameters are searched for both Tf-KD$_{self}$ and Normal KD). For example, with ResNet50 to teach ReseNet18, the student has a 1.19\% improvement, but our method achieves 1.23\% improvement without using any stronger teacher model. We also obtain similar results for MobileNetV2 by Tf-KD$_{self}$ in Fig.~\ref{fig: self_KD_N_KD}.
\vspace{-10pt}

\begin{table}
\centering 
\caption{Accuracy improvement comparison (in $\%$) on CIFAR100 (T: Teacher, R: ResNet, RX: ResNeXt, D: DenseNet).}
\label{tab:self_CIFAR100}
\renewcommand\arraystretch{1.1}
\small
\begin{tabular}{lllll}
\hline 
Model &Baseline & Tf-KD$_{self}$ & Normal KD  [T] \\ \hline
MobileNetV2  &68.38 &70.96 (\textbf{+2.58}) &\textbf{+2.67} [R18]\\ 
ShuffleNetV2 &70.34 &72.23 (\textbf{+1.89}) &\textbf{+1.71}  [R18]\\ 
ResNet18  &75.87 &77.10 (\textbf{+1.23})  &\textbf{+1.19} [R50]\\
GoogLeNet & 78.72 &80.17 (\textbf{+1.45}) &\textbf{+1.39} [RX29]\\ 
DenseNet121 &79.04 &80.26 (\textbf{+1.22}) &\textbf{+1.15} [RX29]\\
ResNeXt29 &81.03 &82.08 (\textbf{+1.05})  &\textbf{+1.12} [RX101]\\ 
\hline  
\end{tabular} 
\vspace{-10pt}
\end{table}

\paragraph{Tiny-ImageNet.} On Tiny-ImageNet, we use baseline models including MobileNetV2, ShuffleNetV2, ResNet50, DenseNet121. They are trained for 200 epochs with batch size $bn=128$ for MobileNetV2, ShuffleNetV2 and $bn=64$ for ResNet50, DenseNet121. The initial learning rate is $\eta =0.1*\frac{bn}{128}$ and then divided by 10 at the 60th, 120th, 160th epoch. We use SGD optimizer with momentum of 0.9, and weight decay is set to 5e-4.
Tab.~\ref{tab:self_t_imagenet} shows the results of Tf-KD$_{self}$ on Tiny-ImageNet. It can be seen that Tf-KD$_{self}$ consistently improves the baseline models and achieves comparable improvement with Normal KD.
\vspace{-10pt}

\begin{table}[!h]
\renewcommand\arraystretch{1.1}
\centering
\caption{Tf-KD$_{self}$ experiment results on Tiny-ImageNet (in \%).}
\small
\label{tab:self_t_imagenet}
\begin{tabular}{lllll}
\hline
Model  & Baseline  & Tf-KD$_{self}$ & Normal KD [T] \\ \hline
MobileNetV2  &55.06 &56.77 (\textbf{+1.71}) &\textbf{+1.64} [R18]\\ 
ShuffleNetV2 &60.51 &61.36 (\textbf{+0.85}) &\textbf{+0.68} [R18]\\ 
ResNet50   &67.47 & 68.18 (\textbf{+0.71}) & \textbf{+0.76} [D121] \\
DenseNet121 &68.15 &68.29 (\textbf{+0.14}) &\textbf{+0.16} [RX29]\\ \hline
\end{tabular}
\vspace{-10pt}
\end{table}

\paragraph{ImageNet.} ImageNet-2012 is one of the largest datasets for object classification, with over 1.3m hand-annotated images. The baseline models we use on this dataset include ResNet18, ResNet50, DenseNet121, RexNeXt101 (32x8d), and we adopt official implementation of Pytorch to train them. We set batch size $bn=512$ for ResNet18, ResNet50, DenseNet121, and $bn=256$ for RexNeXt101. Following common experiment settings~\cite{goyal2017accurate}, the initial learning rate is $\eta =0.1*\frac{bn}{256}$ which is then divided by 10 at the 30th, 60th, 80th epoch in  total 90 epochs. We use SGD optimizer with momentum of 0.9, and weight decay is 1e-4. 
Results are reported in Tab.~\ref{tab:self_imagenet}. We can see that the self-training can further improve the baseline performance on ImageNet-2012. As a comparison, we also use DenseNet121 to teach ResNet18 on ImageNet, and ResNet18 obtains 0.56\% improvement, which is comparable with our Tf-KD$_{self}$ (Tab.~\ref{tab:imagenet_com_with_normalKD}). 
\vspace{-7pt}

\begin{table}[!h]
\renewcommand\arraystretch{1.1}
\centering
\caption{Tf-KD$_{self}$ experiment results on ImageNet (Top1 accuracy, in \%).}
\small
\begin{tabular}{lll}
\hline
Model  & Baseline & Tf-KD$_{self}$  \\ \hline
ResNet18   &69.84 &70.42 (\textbf{+0.58}) \\
ResNet50   &75.77 &76.41 (\textbf{+0.64}) \\
DenseNet121 &75.28 &75.72 (\textbf{+0.44}) \\ 
ResNeXt101 &79.28 &79.56 (\textbf{+0.28}) \\\hline
\end{tabular}
\label{tab:self_imagenet}
\vspace{-15pt}
\end{table}

\begin{table}[!h]
\renewcommand\arraystretch{1.2}
\centering
\caption{Comparison between Tf-KD$_{self}$ and Normal KD on ImageNet (Top1 accuracy, in \%).}
\small
\begin{tabular}{llll}
\hline
Model  & Baseline & Tf-KD$_{self}$ &Normal KD [T] \\ \hline
ResNet18   &69.84 &70.42 (\textbf{+0.58}) & 70.40 (\textbf{+0.56}) [D121]\\\hline
\end{tabular}
\label{tab:imagenet_com_with_normalKD}
\vspace{-10pt}
\end{table}

\begin{table*}[h]
\renewcommand\arraystretch{1.1}
\centering
\caption{Tf-KD$_{reg}$ achieves comparable results with Normal KD on CIFAR100.}
\small
\begin{tabular}{lllll}
\hline 
Model  & Baseline & Tf-KD$_{reg}$ & Normal KD [Teacher] & + LSR\\ \hline
MobileNetV2  &68.38 &70.88 (\textbf{+2.50}) &71.05 (\textbf{+2.67}) [ResNet18] &69.32 (\textbf{+0.94})\\ 
ShuffleNetV2 &70.34 &72.09 (\textbf{+1.75}) &72.05 (\textbf{+1.71}) [ResNet18] &70.83 (\textbf{+0.49})\\ 
ResNet18   &75.87 &77.36 (\textbf{+1.49})  &77.19 (\textbf{+1.32}) [ResNet50] &77.26 (\textbf{+1.39})\\
GoogLeNet  &78.15 &79.22 (\textbf{+1.07}) &78.84 (\textbf{+0.99}) [ResNeXt29] &79.07 (\textbf{+0.92})\\ 
\hline  
\label{tab:reg_CIFAR100}
\end{tabular} 
\vspace{-20pt}
\end{table*}

\begin{table*}[h]
\renewcommand\arraystretch{1.1}
\centering
\caption{Tf-KD$_{reg}$ experiment results on Tiny-ImageNet.}
\small
\label{tab:reg_t_imagenet}
\begin{tabular}{lllll}
\hline
Model  & Baseline & Tf-KD$_{reg}$ & Normal KD [Teacher] & + LSR\\ \hline
MobileNetV2  &55.06 &56.47 (\textbf{+1.41}) & 56.53 (\textbf{+1.47}) [ResNet18] & 56.24 (\textbf{+1.18})\\ 
ShuffleNetV2 &60.51 & 60.93 (\textbf{+0.42}) &61.19 (\textbf{+0.68}) [ResNet18] & 60.66 (\textbf{+0.11})\\ 
ResNet50   &67.47 &67.92 (\textbf{+0.45}) &68.15 (\textbf{+0.68}) [ResNeXt29] &67.63 (\textbf{+0.16)}\\
DenseNet121 &68.15 &68.37 (\textbf{+0.18}) &68.44 (\textbf{+0.26}) [ResNeXt29] &68.19 (\textbf{+0.04})\\ \hline
\end{tabular}
\vspace{-15pt}
\end{table*}

\subsection{Experiments for Manually-designed Regularization}
For all experiments of Tf-KD$_{reg}$, we adopt the same implementation settings with Tf-KD$_{self}$, except for using a virtual output distribution as a regularization term (Eq.~(\ref{eqn:Tf-KD_reg})). For fair comparisons, experiment settings for Normal KD and Tf-KD$_{reg}$ are the same. See Supplementary Material for hyper-parameters of Tf-KD$_{reg}$.
\vspace{-10pt}

\paragraph{CIFAR100 and Tiny-ImageNet.}  
For Tf-KD$_{reg}$ experiments on CIFAR100 and Tiny-ImageNet, we set the probability for correct classes as $a=0.99$ (Eq.~(\ref{eqn:T_probability})). The temperature $\tau$ and $\alpha$ in Eq.~(\ref{eqn:Tf-KD_reg}) are different for different baseline models (see Supplementary Material). From Tab.~\ref{tab:reg_CIFAR100} and Tab.~\ref{tab:reg_t_imagenet}, we can observe with no teacher used and just a regularization term added, Tf-KD$_{reg}$ achieves comparable performance with Normal KD on both CIFAR100 and Tiny-ImageNet. 
\vspace{-10pt}

\paragraph{ImageNet.} For the Tf-KD$_{reg}$ on ImageNet, we adopt temperature $\tau=20$ as normal knowledge distillation, and $\alpha=0.1$ as label smoothing regularization. The probability for correct classes in the manually-designed teacher is $a=0.99$ (Eq.~(\ref{eqn:Tf-KD_reg})). We test our Tf-KD$_{reg}$ with four baseline models: ResNet18, ResNet50, DenseNet121 and ResNeXt101 (32x8d). As a regularization term, the manually designed teacher achieves consistent improvement compared with baselines. For example, the proposed Tf-KD$_{reg}$ improves the top1 accuracy of ResNet50 by 0.65\% on ImageNet-2012 (Tab.~\ref{tab:KD_man_imagenet}). Even for a huge single model ResNeXt101 (32x8d) with 88.79M parameters, our method achieves 0.48\% improvement by using the manually designed teacher. 
\vspace{-10pt}

\begin{table}[h]
\centering
\caption{Test accuracy improvement (in $\%$) on ImageNet.}
\begin{tabular}{llll}
\hline
Model  & Baseline & +Tf-KD$_{reg}$  & + LSR\\ \hline
ResNet18   &69.84 &70.24 (\textbf{+0.40}) &70.02 (\textbf{+0.18})\\
ResNet50   &75.77 &76.42 (\textbf{+0.65}) &76.38 (\textbf{+0.51}) \\
DenseNet121 &75.28 &75.62 (\textbf{+0.34}) &75.24 (\textbf{-0.04}) \\ 
ResNeXt101 &79.28 &79.76 (\textbf{+0.48}) &79.67 (\textbf{+0.39})\\\hline
\end{tabular}
\label{tab:KD_man_imagenet}
\vspace{-10pt}
\end{table}

Comparing our two methods Tf-KD$_{self}$ and Tf-KD$_{reg}$, we observe that Tf-KD$_{self}$ works better in small dataset (CIFAR100) while Tf-KD$_{reg}$ performs slightly better in large dataset (ImageNet).

\vspace{-15pt}
\paragraph{Comparison with LSR} The Tf-KD$_{reg}$ is motivated by LSR, which can be seen as a modification of LSR. This modification significantly improves the performance of neuron networks without extra computation cost. Same as LSR, Tf-KD$_{reg}$ can serve as a generic regularization method to normally train neural networks. We compare our Tf-KD$_{reg}$ with label smoothing on CIFAR100, Tiny-ImageNet and ImageNet. For fair comparisons, experiment settings for Tf-KD$_{reg}$ and LSR are the same. The results are shown in Tab.~\ref{tab:reg_CIFAR100},~\ref{tab:reg_t_imagenet} and~\ref{tab:KD_man_imagenet}. It can be seen that Tf-KD$_{reg}$ consistently outperforms LSR. Additionally, the formulation of KDR$_{man}$ is similar to LSR, but it is not an over-parameterized version of label smoothing. We give detailed comparison between Tf-KD$_{reg}$ and LSR to show the difference in Supplementary Material.

\section{Related Work}
\vspace{-5pt}
\paragraph{Knowledge Distillation} Since \cite{hinton2015distilling} proposed knowledge distillation based on prior work~\cite{ba2014deep}, KD has been widely adopted or modified~\cite{romero2014fitnets,yim2017gift,yu2017visual,furlanello2018born, anil2018large, mirzadeh2019improved, Wang_2019_CVPR}. Different from existing works, our work challenges the common belief of knowledge distillation based on our designed exploratory experiments. A related work is deep mutual learning~\cite{zhang2018deep}, which proposes to let an ensemble of student models to learn with each other by minimizing the KL Divergence of predictions. Comparatively, our work reveals the relationship between KD and label smoothing, and our proposed Tf-KD can serve as a general method for neural network training. Another related work is Born-again networks~\cite{furlanello2018born}, which use similar method as Tf-KD$_{self}$. The difference is that Born-again networks utilize an ensemble of students to train itself in the final step. 

\vspace{-14pt}
\paragraph{Label Smoothing} Szegedy et al.~\cite{szegedy2016rethinking} proposed LSR to replace the ``hard labels'' with smoothed labels, boosting performance of many tasks like image classification, language translation and speech recognition~\cite{pereyra2017regularizing}. Recently, \cite{muller2019does} empirically showed label smoothing can also help improve model calibration. In our work, we adopt label smoothing regularization to understand the regularization function of knowledge distillation. 

\section{Conclusion}
In this work, we find through experiments and analyses that the ``dark knowledge'' of a teacher model is more of a regularization term than similarity information of categories. Based on the relationship between KD and LSR, we propose Teacher-free KD. Experiment results show our Tf-KD can achieve comparable results with Normal KD in image classification. 
Our work also suggests that, when it is hard to find a stronger teacher for a powerful model or computation resource is limited to train teacher models, the targeted model can still get enhanced by self-training or a manually-designed regularization term. 
\paragraph{Acknowledgement} Jiashi Feng was partially supported by AI.SG R-263-000-D97-490, NUS ECRA R-263-000-C87-133 and MOE Tier-II R-263-000-D17-112. Besides, we thank Dr.Jianan Li, Mr.Daquan Zhou and Mr.Yujun Shi for discussion during this work. 


{\small
\bibliographystyle{ieee}
\bibliography{egbib}

\begin{thebibliography}{10}\itemsep=-1pt

\bibitem{anil2018large}
R.~Anil, G.~Pereyra, A.~Passos, R.~Ormandi, G.~E. Dahl, and G.~E. Hinton.
\newblock Large scale distributed neural network training through online
  distillation.
\newblock {\em arXiv preprint arXiv:1804.03235}, 2018.

\bibitem{ba2014deep}
J.~Ba and R.~Caruana.
\newblock Do deep nets really need to be deep?
\newblock In {\em Advances in neural information processing systems}, pages
  2654--2662, 2014.

\bibitem{deng2009imagenet}
J.~Deng, W.~Dong, R.~Socher, L.-J. Li, K.~Li, and L.~Fei-Fei.
\newblock Imagenet: A large-scale hierarchical image database.
\newblock In {\em 2009 IEEE conference on computer vision and pattern
  recognition}, pages 248--255. Ieee, 2009.

\bibitem{furlanello2018born}
T.~Furlanello, Z.~C. Lipton, M.~Tschannen, L.~Itti, and A.~Anandkumar.
\newblock Born again neural networks.
\newblock {\em arXiv preprint arXiv:1805.04770}, 2018.

\bibitem{goyal2017accurate}
P.~Goyal, P.~Doll{\'a}r, R.~Girshick, P.~Noordhuis, L.~Wesolowski, A.~Kyrola,
  A.~Tulloch, Y.~Jia, and K.~He.
\newblock Accurate, large minibatch sgd: Training imagenet in 1 hour.
\newblock {\em arXiv preprint arXiv:1706.02677}, 2017.

\bibitem{he2016deep}
K.~He, X.~Zhang, S.~Ren, and J.~Sun.
\newblock Deep residual learning for image recognition.
\newblock In {\em Proceedings of the IEEE conference on computer vision and
  pattern recognition}, pages 770--778, 2016.

\bibitem{hinton2015distilling}
G.~Hinton, O.~Vinyals, and J.~Dean.
\newblock Distilling the knowledge in a neural network.
\newblock {\em arXiv preprint arXiv:1503.02531}, 2015.

\bibitem{huang2017densely}
G.~Huang, Z.~Liu, L.~Van Der~Maaten, and K.~Q. Weinberger.
\newblock Densely connected convolutional networks.
\newblock In {\em Proceedings of the IEEE conference on computer vision and
  pattern recognition}, pages 4700--4708, 2017.

\bibitem{krizhevsky2009learning}
A.~Krizhevsky, G.~Hinton, et~al.
\newblock Learning multiple layers of features from tiny images.
\newblock Technical report, Citeseer, 2009.

\bibitem{ma2018shufflenet}
N.~Ma, X.~Zhang, H.-T. Zheng, and J.~Sun.
\newblock Shufflenet v2: Practical guidelines for efficient cnn architecture
  design.
\newblock In {\em Proceedings of the European Conference on Computer Vision
  (ECCV)}, pages 116--131, 2018.

\bibitem{mirzadeh2019improved}
S.-I. Mirzadeh, M.~Farajtabar, A.~Li, and H.~Ghasemzadeh.
\newblock Improved knowledge distillation via teacher assistant: Bridging the
  gap between student and teacher.
\newblock {\em arXiv preprint arXiv:1902.03393}, 2019.

\bibitem{muller2019does}
R.~M{\"u}ller, S.~Kornblith, and G.~Hinton.
\newblock When does label smoothing help?
\newblock {\em arXiv preprint arXiv:1906.02629}, 2019.

\bibitem{pereyra2017regularizing}
G.~Pereyra, G.~Tucker, J.~Chorowski, {\L}.~Kaiser, and G.~Hinton.
\newblock Regularizing neural networks by penalizing confident output
  distributions.
\newblock {\em arXiv preprint arXiv:1701.06548}, 2017.

\bibitem{romero2014fitnets}
A.~Romero, N.~Ballas, S.~E. Kahou, A.~Chassang, C.~Gatta, and Y.~Bengio.
\newblock Fitnets: Hints for thin deep nets.
\newblock {\em arXiv preprint arXiv:1412.6550}, 2014.

\bibitem{sandler2018mobilenetv2}
M.~Sandler, A.~Howard, M.~Zhu, A.~Zhmoginov, and L.-C. Chen.
\newblock Mobilenetv2: Inverted residuals and linear bottlenecks.
\newblock In {\em Proceedings of the IEEE Conference on Computer Vision and
  Pattern Recognition}, pages 4510--4520, 2018.

\bibitem{szegedy2016rethinking}
C.~Szegedy, V.~Vanhoucke, S.~Ioffe, J.~Shlens, and Z.~Wojna.
\newblock Rethinking the inception architecture for computer vision.
\newblock In {\em Proceedings of the IEEE conference on computer vision and
  pattern recognition}, pages 2818--2826, 2016.

\bibitem{Wang_2019_CVPR}
T.~Wang, L.~Yuan, X.~Zhang, and J.~Feng.
\newblock Distilling object detectors with fine-grained feature imitation.
\newblock In {\em The IEEE Conference on Computer Vision and Pattern
  Recognition (CVPR)}, June 2019.

\bibitem{xie2017aggregated}
S.~Xie, R.~Girshick, P.~Doll{\'a}r, Z.~Tu, and K.~He.
\newblock Aggregated residual transformations for deep neural networks.
\newblock In {\em Proceedings of the IEEE conference on computer vision and
  pattern recognition}, pages 1492--1500, 2017.

\bibitem{yim2017gift}
J.~Yim, D.~Joo, J.~Bae, and J.~Kim.
\newblock A gift from knowledge distillation: Fast optimization, network
  minimization and transfer learning.
\newblock In {\em Proceedings of the IEEE Conference on Computer Vision and
  Pattern Recognition}, pages 4133--4141, 2017.

\bibitem{yu2017visual}
R.~Yu, A.~Li, V.~I. Morariu, and L.~S. Davis.
\newblock Visual relationship detection with internal and external linguistic
  knowledge distillation.
\newblock In {\em Proceedings of the IEEE International Conference on Computer
  Vision}, pages 1974--1982, 2017.

\bibitem{zhang2018deep}
Y.~Zhang, T.~Xiang, T.~M. Hospedales, and H.~Lu.
\newblock Deep mutual learning.
\newblock In {\em Proceedings of the IEEE Conference on Computer Vision and
  Pattern Recognition}, pages 4320--4328, 2018.

\end{thebibliography}
}

\end{document}